\documentclass[letterpaper, 10 pt, conference]{ieeeconf}  

\IEEEoverridecommandlockouts       

\usepackage[backend=biber,
            url=false,
            isbn=false,
            doi=false,
            backref=false,
            style=ieee,
            natbib=true,
            mincitenames=1,
            maxcitenames=1,
            citestyle=numeric-verb,
            sorting=none,
            block=none]{biblatex}

\renewcommand{\bibfont}{\small}

\usepackage{graphics}
\usepackage[pdftex]{graphicx}
\usepackage{wrapfig}
\DeclareGraphicsExtensions{.pdf,.png,.jpg}
\usepackage{epsfig}
\usepackage[font={small}]{caption}
\usepackage{subcaption}
\usepackage[rightcaption]{sidecap}
\usepackage{pbox}
\usepackage{bigstrut}
\usepackage{makecell}
\usepackage{ctable} 

\usepackage{amssymb,amsmath}
\usepackage{gensymb} 
\usepackage{nicefrac}       
\numberwithin{equation}{section} 
\usepackage{algorithm}
\usepackage{algpseudocode}

\usepackage{textcomp} 

\usepackage{array} 
\usepackage{tabularx}
\usepackage{multirow}
\usepackage{multicol}
\usepackage{booktabs}
\usepackage{tabulary}

\usepackage[utf8]{inputenc}
\usepackage{units}
\usepackage{bm}
\usepackage{xspace}
\usepackage{flushend}
\usepackage{balance} 
\usepackage{csquotes}
\usepackage{makeidx}
\usepackage{blindtext}
\usepackage{xcolor}
\usepackage{caption}
\usepackage{subcaption}

\usepackage{url}

\usepackage{hyperref}
\hypersetup{
    colorlinks=true,
    linkcolor=black,
    citecolor=black,
    filecolor=cyan,
    urlcolor=black
}

\renewcommand{\baselinestretch}{1.0}



\title{\LARGE \textbf{
Automating Vascular Shunt Insertion with the dVRK Surgical Robot}}

\author{
Karthik Dharmarajan$^1$*, Will Panitch$^1$*, Muyan Jiang$^1$, Kishore Srinivas$^1$, Baiyu Shi$^1$\\
Yahav Avigal$^1$, Huang Huang$^1$, Thomas Low$^2$, Danyal Fer$^3$, Ken Goldberg$^1$\\
\thanks{* equal contribution}
\thanks{$^1$ The AUTOLab at UC Berkeley (\href{mailto:automation@berkeley.edu}{automation@berkeley.edu})}
\thanks{$^2$ SRI International}
\thanks{$^3$ UC San Francisco Medical School}
}
\begin{document}

\maketitle
\begin{abstract}
Vascular shunt insertion is a fundamental surgical procedure used to temporarily restore blood flow to tissues. It is often performed in the field after major trauma. We formulate a problem of automated vascular shunt insertion and propose a pipeline to perform Automated Vascular Shunt Insertion (AVSI) using a da Vinci Research Kit. The pipeline uses a learned visual model to estimate the locus of the vessel rim, plans a grasp on the rim, and moves to grasp at that point. The first robot gripper then pulls the rim to stretch open the vessel with a dilation motion. The second robot gripper then proceeds to insert a shunt into the vessel phantom (a model of the blood vessel) with a chamfer tilt followed by a screw motion. Results suggest that AVSI achieves a high success rate even with tight tolerances and varying vessel orientations up to 30\degree. Supplementary material, dataset, videos, and visualizations can be found at \url{https://sites.google.com/berkeley.edu/autolab-avsi}. 
\end{abstract}

\section{Introduction}

\begin{figure}
\includegraphics[width=0.4\textwidth]{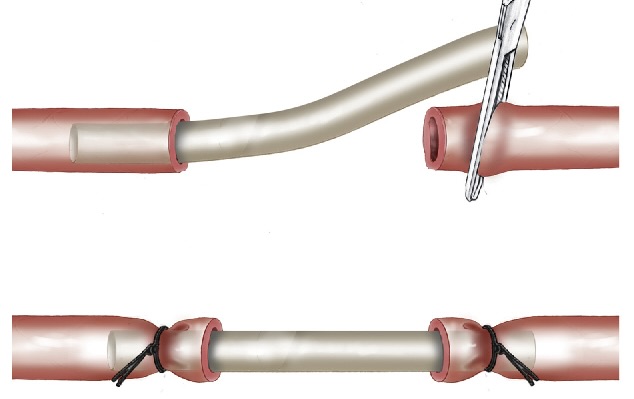}
\centering
\caption{\textbf{Vascular shunt Insertion.} The lumen of a blood vessel is widened for the insertion of a vascular shunt. Two ligature knots are applied to maintain the position of the shunt afterwards. \citep{Voiglio2016AbbreviatedLO}}
\label{fig:shuntprocedure}
    \vspace{5pt}
\end{figure}

A shunt is a hollow, flexible catheter that is placed within the human body in order to divert fluid from one location to another~\citep{cannon_2018}. Vascular shunts, which are placed to bridge gaps left by blood vessel trauma (Fig. \ref{fig:shuntprocedure}), are widely utilized in civilian and battlefield settings to quickly restore blood flow to areas which might otherwise be at risk due to vascular damage~\citep{subramanian2008decade,rasmussen2006use}. Due to their lifesaving and time-sensitive nature, vascular shunt procedures often take place in chaotic or high-stakes clinical scenarios in which precision is critical but surgeon fatigue and disruptions are commonplace. To improve the consistency and effectiveness of this procedure, we develop a reliable and efficient policy for performing autonomous vascular shunt insertion utilizing computer vision and Robotic Surgical Assistants (RSAs).

Techniques based on computer vision approaches have been used in ventriculoperitoneal shunt settings for 3D ventricle segmentation and computation of ventricular shunt-placement locations~\citep{liu2021clinical, doddamani2021robot}. In these cases, such techniques served as an aid for human surgeons during preoperative stage, rather than as part of a purely autonomous pipeline. Furthermore, to the best of our knowledge, computer vision methods have not previously been applied to shunt insertion in the vascular setting. RSAs have been used in phantom vascular shunt operations~\cite{garcia2009trauma}; however, these systems were under complete control of the operating surgeon at all times. Our problem formulation is distinct, as it requires the robot to perform the dilation and insertion procedure without direct human involvement. In the fully-robotic setting, this procedure presents challenges due to self-occlusion of the end effector and vessel rim, as well as the deformable object manipulation required to grasp and dilate the vessel with an appropriate amount of force. Furthermore, the millimeter-level precision that this task demands is difficult for cable-driven RSAs to achieve, given their propensity for slippage and hysteresis.
\begin{figure}[t!]
\centering
\includegraphics[width=1.0\linewidth]{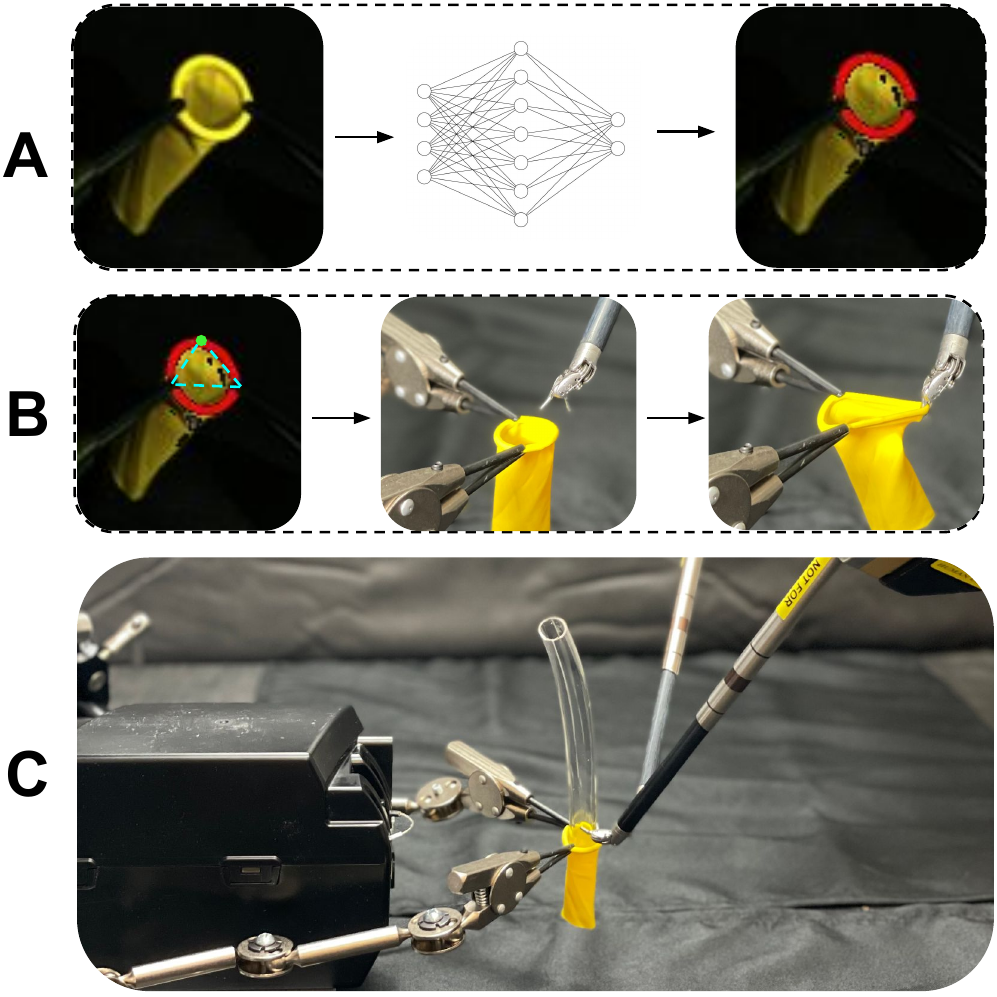}
\caption{
\textbf{Overview of pipeline}:  
\textbf{A)} Given a stereo RGB input from the camera, the pipeline produces a segmentation mask and a fitted circle of the rim of the vessel phantom using a pre-trained neural network. 
\textbf{B)} The robot then computes an equidistant grasp point from the two fixed points on the rim, plans a motion path, moves the gripper to the rim, and pulls the rim outward to dilate it. 
\textbf{C)} After the dilation, the second robot gripper inserts a shunt into the vessel phantom using chamfer tilting and screw motion.}
\vspace{5pt}
\label{fig:figure_one}
\end{figure}

In this paper we formalize a specific vascular shunt insertion procedure and propose an algorithm using computer vision, active sensing, and deep learning. Given a vessel phantom mounted at two points along its rim, our procedure identifies the rim of the vessel, computes a grasp point along the rim that is equally far away from both fixed points, applies deep calibration to execute an accurate grasp, pulls outward to apply tension at this grasp point, and inserts a shunt into the vessel phantom.

We test our system's ability to dilate and insert shunts into vessel phantoms of varying initial orientations and shunt sizes in different runs consisting of 20 trials each.

This paper contributes:
\begin{enumerate}
    \item A formulation for the AVSI problem.
    \item An algorithm for AVSI with a da Vinci Research Kit Robotic Surgical Assistant.
    \item Physical experimental results that suggest a success rate of between 80\% and 95\% and an average completion time of between 13.7s and 14.4s.
    \item A dataset of 1500 RGB images and corresponding segmentation masks of a vessel phantom at different poses. 
\end{enumerate}
\section{Related Work}\label{sec:related_work}

\subsection{Vascular Shunt Insertion}
After a trauma, to salvage a limb, blood flow generally needs to be restored within 6-8 hours after injury~\citep{subramanian2008decade, eger1971use, malan1963physio} and animal model data indicates that restoration of flow in less than three hours has improved outcomes~\citep{burkhardt2010large}. The development of plastic shunts emerged as a solution to quickly restore blood flow in the trauma setting, particularly when the surgeon did not have the skill set or the time to perform a definitive vascular repair. Vascular shunts have entered significant utilization in more recent conflicts where “damage control” surgical care is occurring near the front lines to perfuse a limb prior to transport or allow for additional resuscitation in the intensive care unit~\citep{rasmussen2006echelons}. Ultimately these shunts will be removed within 12-48 hours when a more definitive vascular repair will be performed. 



\subsection{Automation in Robot-Assisted Surgery}

Automation of surgical subtasks in laboratory settings is an active area of research. Several subtasks have already been studied in previous literature such as debridment \citep{Kehoe2014,seita_icra_2018}, peg transfer \citep{auto_peg_transfer_2015, hwang2020applying, hwang2020efficiently, paradis2020intermittent, }, surgical cutting \citep{krishnan2019swirl, murali2015learning, thananjeyan2017multilateral}, cutting gauze \citep{thananjeyan2017multilateral, rosen_icra_tissues_2019} and suturing \citep{sen2016automating, automated_needle_pickup_2018, saeidi_suturing_icra_2019,  wilcox2021learning}. 

There has been a recent breakthrough in autonomous robotic surgery in the laparoscopic setting for intestinal anastomosis \citep{saeidi2022autonomous}, which has successfully performed expert surgeons' technique. However, this has yet to be generalized to location-agnostic shunt procedures. To the best of our knowledge, fully automated procedures for general vessel dilation or shunt insertion have not been explored yet.

\subsection{Vessel Rim Locus Detection}

The use of convolutional neural networks (CNNs) is a popular technique in feature detection within surgical video imaging \citep{bamba2021object, sanchez2022gauze}, and has mainly been applied to surgical tool detection \citep{yang2020image, bamba2021automated, liu2022real, boonkong2022surgical}. Other applications include the identification of laryngeal nerves during thyroidectomy \citep{gong2021using}, tumour-targeting \citep{goto2022image}, and other organ anomalies \citep{chheda2020gastrointestinal}. Most of these involve some variation of regional convolutional neural networks (R-CNNs), the YOLO network, or a transformer-CNN cross model to form the object detection in specific lighting conditions. More specifically, there exist methods for detecting circular holes (similar to those of vessel openings) in industrial settings \citep{li2021semi, ling4075794deep, prabuwono2019automated} as well as pupil edge extraction for cataract surgeries \citep{ektesabi2011exact}. 

To the best of our knowledge, this specific vessel rim localization problem has not previously been attempted.

\section{Problem Statement}\label{sec:ps}

\subsection{Overview}
We consider the case of a bimanual surgical robot which dilates the opening of an unknown vessel phantom and inserts a known shunt. Specifically, we assume that the rim of the vessel is held by two fixed grippers which resemble the role played by the surgical assistant in a traditional vascular shunt operation. We further assume that the robot is grasping the shunt in a known pose, and an RGBD camera is mounted above the workspace. We also assume the rigid transformations between the camera, robot, and workspace coordinate frames are known.
We assume the RGBD camera provides an RGB image and depth image during each iteration, the radius of the shunt is less than or equal to the radius of the undilated vessel, and the vessel rim maintains a circular shape to best resemble the opening of a real blood vessel.

\subsection{Objective and Evaluation Metrics}
A trial consists of the robot gripping the identified grasp point on the vessel rim, stretching the vessel, and inserting a shunt into the vessel.
We consider a trial successful if the rim of the shunt is fully enclosed by the vessel after both grippers release their grips and retract back to their original position. Note that this is only possible if both the grasp and dilation steps are successful, so we do not consider separate success metrics for these steps. The result of a successful trial is shown in Fig. \ref{fig:shuntinsertion}. 

\section{Method}\label{sec:method}
As illustrated in Fig. \ref{fig:figure_one}, our pipeline consists of three main stages: vessel phantom rim state estimation, vessel phantom rim dilation, and shunt insertion.

\begin{figure}[t!]
     \centering
     \begin{tabular}{c}
         \begin{subfigure}[b]{0.155\textwidth}
             \centering
             \includegraphics[width=\textwidth]{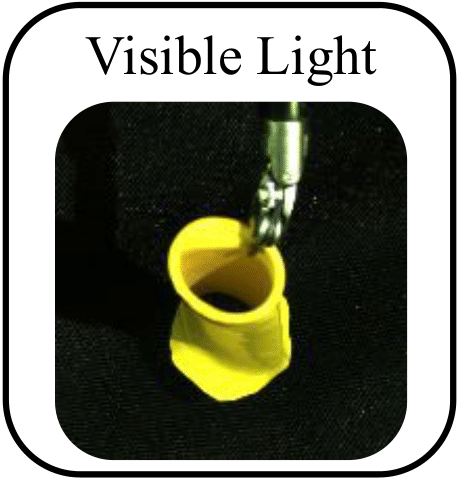}
             \caption{}
         \end{subfigure}
         \hfill
         \begin{subfigure}[b]{0.155\textwidth}
             \centering
             \includegraphics[width=\textwidth]{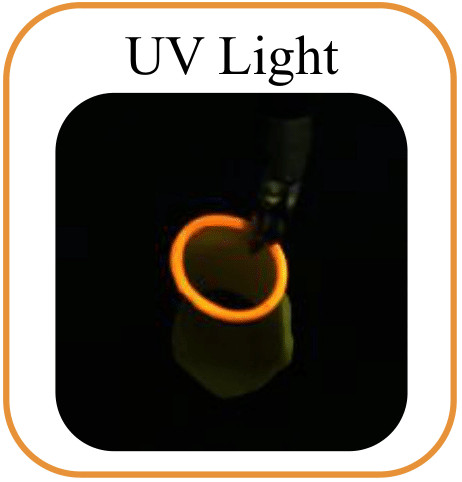}
             \caption{}
         \end{subfigure}
         \hfill
         \begin{subfigure}[b]{0.155\textwidth}
             \centering
             \includegraphics[width=\textwidth]{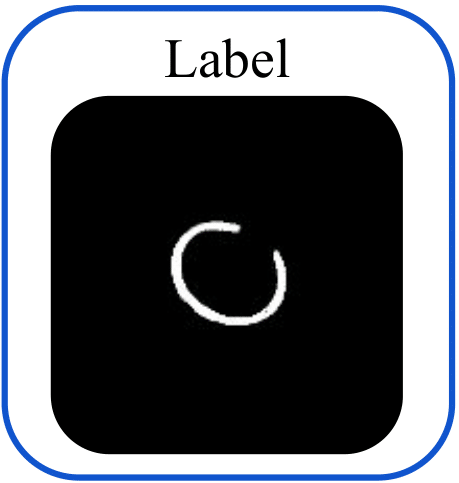}
             \caption{}
         \end{subfigure} \\
         \begin{subfigure}[b]{0.18\textwidth}
             \centering
             \includegraphics[width=\columnwidth]{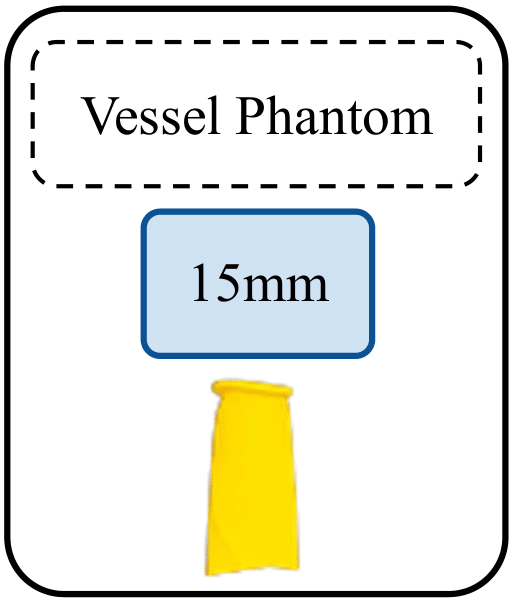}
             \caption{}
         \end{subfigure}
         \hfill
         \begin{subfigure}[b]{0.285\textwidth}
             \centering
             \includegraphics[width=\columnwidth]{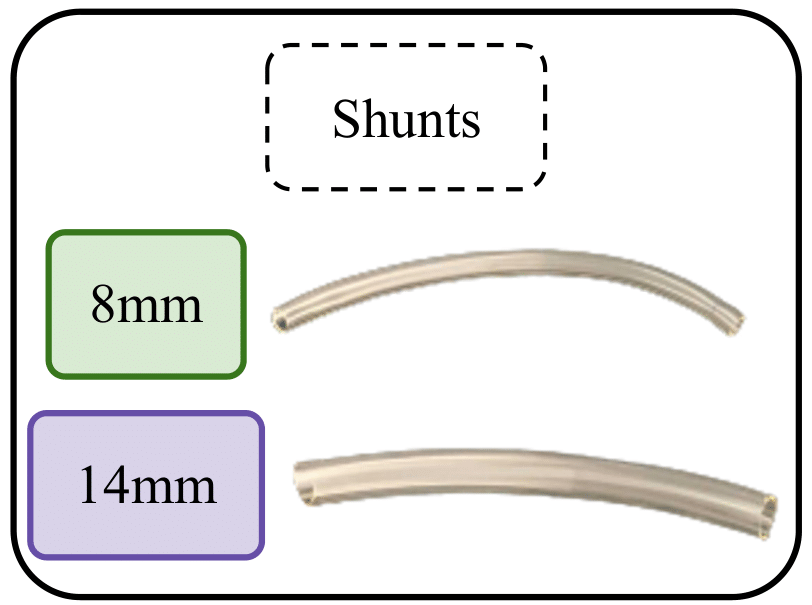}
             \caption{}
         \end{subfigure}
     \end{tabular}
     \caption{\textbf{Data collection and experiment objects.} 
\textbf{Top:} We collect data for training our neural network, consisting of a) an image of the vessel phantom when exposed to visible light, b) an image of the phantom vessel when exposed to UV light, and c) the extracted segmentation mask. \textbf{Bottom:} We use d)  a 15mm inner diameter vessel phantom (left, yellow) and e) two shunts (right, translucent) with 8mm and 14mm outer diameters.}
    \label{fig:trainingdata}
        \vspace{7pt}
\end{figure}

\subsection{Vessel Phantom Rim State Estimation}\label{subsec:rim_state_estimation}
In the first step of the procedure, the algorithm determines the state of the rim of the vessel phantom from an RGBD image. The shape and orientation of the vessel phantom rim are characterized as a circle in Cartesian space, consisting of a center point $c_p$, normal vector $c_n$, and radius $r$ (Fig.~\ref{fig:scene}). Estimating this circle from an RGBD image consists of two stages: \emph{Vessel Phantom Rim Segmentation} and \emph{Circle Fitting}.

\subsubsection{Vessel Phantom Rim Segmentation}\label{subsec:rim_segmentation}

Vessel phantom rim segmentation converts an RGB image of the workspace to a segmentation mask marking the location of the rim of the vessel. We train an asymmetric U-Net~\citep{ronneberger2015u, asymmetrictumor} to generate the segmentation masks (Fig. \ref{fig:trainingdata}(c)) from RGB images (Fig. \ref{fig:trainingdata}(a)) using LUV \cite{thananjeyan2022all}, an unsupervised label collection technique. By painting the top of the vessel phantom with ultraviolet fluorescent paint and utilizing a remotely-controllable UV and visible lighting system, we are able to quickly collect paired visible light and UV light (Fig. \ref{fig:trainingdata}(b)) images, in which the fluorescent glow allows for the extraction of segmentation masks for training via color thresholding. This data collection method is not specific to a particular vessel phantom, but can be extended to other vessel phantoms or vessels. As in \cite{thananjeyan2022all}, we utilize a network architecture with a 3-tier contracting path and a 3-tier expansive path. However, by replacing the final "up-convolution" level of the expansive path with an upsampling layer, we reduce the network runtime and parameter count. The segmentation mask is then projected onto the point cloud generated by the RGBD camera to select the points on the vessel rim in 3D space. 

\begin{figure}[t!]
     \centering
     \begin{tabular}{c}
        \includegraphics[width=.93\columnwidth]{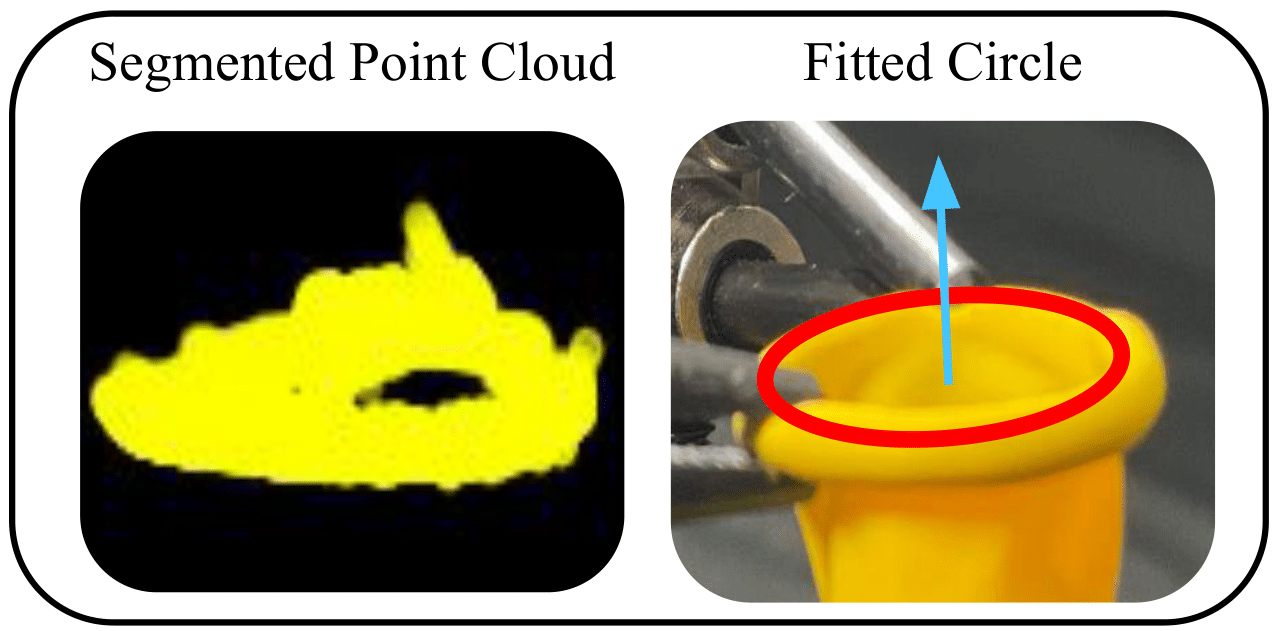} \\
     \end{tabular}
     \caption{\textbf{Circle fitting.} 
\textbf{Left:} Side view of a segmented noisy point cloud deprojected from the RGBD image. \textbf{Right:} We use RANSAC to fit a circle (in red) with its normal vector (in blue) and both elements on a side view of the same scene of the point cloud.}
    \label{fig:circle_fitting}
    \vspace{7pt}
\end{figure}

We collected $3200$ UV/visible light image pairs for training across a variety of different vessel phantom materials and sizes, as well as under different gripper orientations and workspace layouts. A further $805$ image pairs with corresponding 3D depth data were held out as a validation set to test the performance of our full localization pipeline. In addition, to ensure that our perception system did not become overly reliant on the color profile of our vessel phantoms (which could differ significantly from the conditions in an in-vivo operation), we additionally trained an otherwise-identical version of our perception pipeline using only grayscale input information for comparison.



\subsubsection{Circle Fitting}

Due to the abundance of outliers in the noisy point cloud and neural network segmentation mask output, we apply random sample consensus (RANSAC)~\cite{fischler1981random} to estimate the state of the rim represented as a tuple $(c_p, c_n, r)$. At each iteration of the RANSAC algorithm, we sample 3 points from the point cloud, to which we fit a circle. The best fit circle seen so far is kept and returned at the end of the algorithm (Fig.~\ref{fig:circle_fitting}). Our pipeline uses a RANSAC inlier radius of 1mm and runs for 1000 iterations, parameters based upon empirical performance. 

\subsection{Vessel Phantom Rim Dilation}
In the second part of the pipeline, after the rim of the vessel phantom is estimated, the algorithm plans a grasp and moves to that grasping point in a manner that avoids collision with the vessel phantom.

\begin{figure*}
\includegraphics[width=0.95\textwidth]{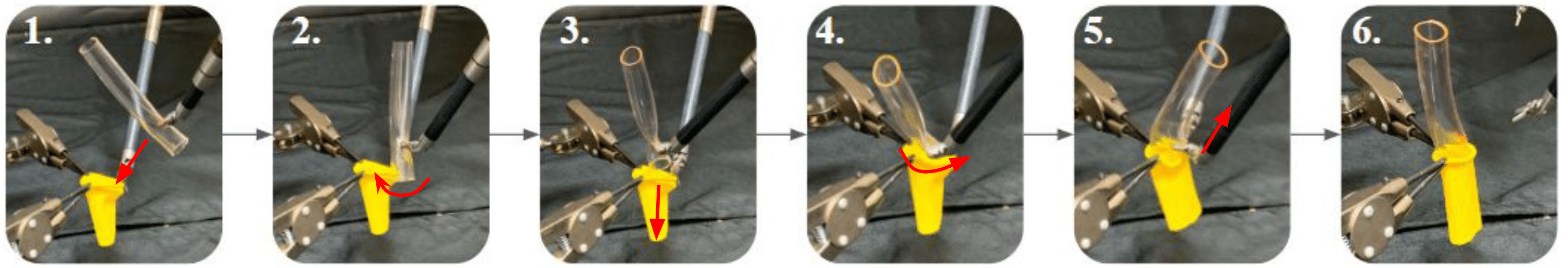}
\centering
\captionsetup{width=0.95\textwidth}
\caption{\textbf{Chamfer tilt shunt insertion with screw motion.} 1) One gripper dilates the vessel. 2) The other gripper, with the shunt already held, moves to a position outside of the rim of the phantom vessel. 3) The second gripper then moves above the vessel phantom while tilting the shunt. 4) The gripper moves downward, resulting in a part of the shunt being contained within the phantom vessel. 5) The gripper straightens the tilt and executes a screw motion, rotating counterclockwise and moving downward. While this is happening, the gripper dilating the phantom vessel moves up and towards the center of the vessel phantom, providing a slight release of tension. 6) Finally, both grippers release and return to starting positions, leaving the shunt inside of the phantom vessel. Red arrows indicate the direction of motion starting from that state to go to the next state.}
\label{fig:shuntinsertion}
\end{figure*}
\subsubsection{Vessel Phantom Grasp Planning}

\begin{figure}[h]
\centering
\frame{\includegraphics[width=.7\columnwidth]{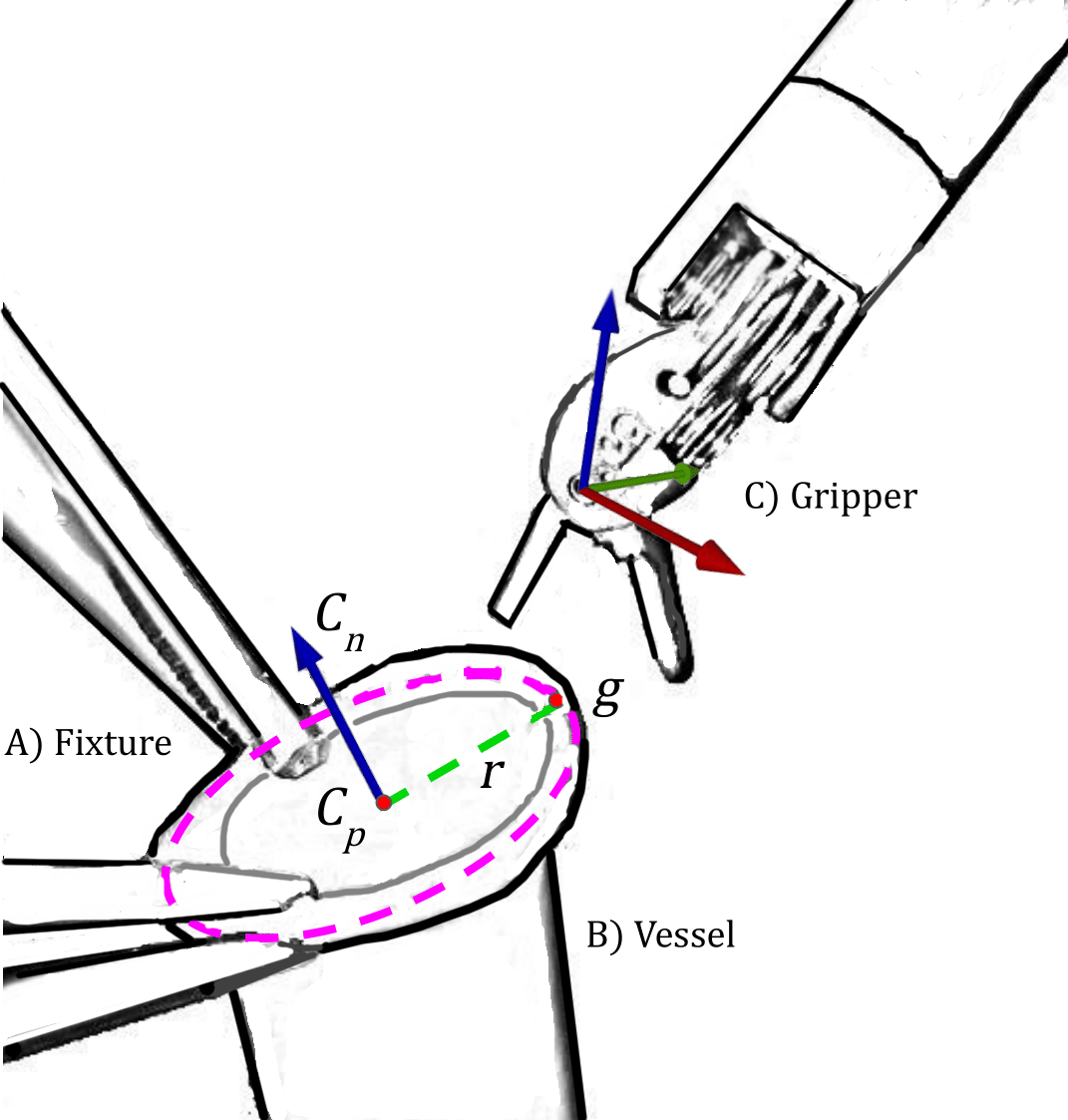}}
\caption{\textbf{Grasp Planning.}  After the segmentation and circle fitting, we output a circle (in purple) represented by its center, normal vector and radius $(c_p,c_n,r)$. Grasping point $g$ is computed to be the equidistant point on the circle to the fixed points. The coordinate frame on the end effector with x-y-z axis corresponding to R-G-B color respectively is rotated in such a way that z-axis aligns with $c_n$, x-axis aligns with $g-c_p$ and y-axis aligns with the cross product of the two. }
\label{fig:scene}
    \vspace{5pt}
\end{figure}

The algorithm computes a grasp point, $g$, on the estimated circle and that is equidistant from the two fixed points. For the orientation of the end effector (Fig.~\ref{fig:scene}), the z-axis is set as the normal vector $c_n$ of the circle, the x-axis is set as a unit vector in the direction of $g - c_p$, and the y-axis is set as the cross product of z and x.

Moving directly to the grasp point can cause collisions with the vessel phantom. To address this issue, the arm first moves to a point 5mm away from $g$ in the direction of the fit circle's normal vector. The 5mm distance allows an open gripper to remain collision free when moving from its start pose towards the grasp point, $g$. Then, the arm moves toward the rim in the direction of $-c_n$ and then grasps the rim.

\subsubsection{Vessel Phantom Dilation}
To dilate the vessel phantom, the algorithm computes a point for the final position of the end effector by starting at the currently grasped point and moving outward from the center of the circle by a distance that is $\frac{2}{3}$ of the diameter of the fit circle. The dilation movement of $\frac{2}{3}$ of the diameter is chosen based upon physical observations of the vessel phantom to balance the gain of additional area in the opening and the stress placed upon it. To reduce the probability of the end effector losing grip on the vessel phantom and to reduce the risk of tearing the vessel, the speed of the outward motion is limited to 25\% of the robot's maximum velocity. We do not incorporate end effector forces directly into the motion, as accurately measuring them at the end effector is challenging.

\subsection{Shunt Insertion}
After vessel phantom dilation, the shape of the rim becomes triangular (Fig. \ref{fig:figure_one}). The gripper used to tension the vessel now becomes an obstacle that the gripper holding the shunt must avoid when inserting. Furthermore, when the outer diameter of the shunt is close to the inner diameter of the vessel phantom, the shunt creates a very tight fit. To overcome the challenges associated with the tight fit, the method utilizes an initial chamfer tilt insertion followed by a screw motion to insert tightly fitting shunts.

\subsubsection{Chamfer Tilt Insertion}

To first make it easier to insert the shunt and set up the screw motion, the robot moves toward a point slightly outside of the vessel phantom rim. It then proceeds to move above the vessel phantom, and it rotates the end effector such that the shunt is tilted with respect to the rim. The end effector moves downward, inserting part of the tilted shunt into the vessel phantom. Then, the end effector rotates such that the shunt is no longer tilted. To further improve the fit of the vessel around the shunt, the arm dilating the vessel moves upward and towards the center. The detailed motion is summarized in Fig. \ref{fig:shuntinsertion} by actions 1 to 5.

\subsubsection{Screw Motion}

In certain cases, where there is a small part of the shunt still outside of the rim of the vessel, after both grippers release the vessel, the shunt may not stay in place. To increase the probability that the entire end of the shunt is within the vessel phantom, the arm executes a screw motion, a combination of counterclockwise rotation and downwards translation. The screw motion makes any part of the shunt already inside of the vessel phantom stay inside of the vessel phantom, while providing an opportunity for parts outside of the vessel phantom to move inside.
\section{Experiments}\label{sec:experiments}
\begin{table*}[h]
\centering
\resizebox{\textwidth}{!}{
 \begin{tabular}{ c c c c c c c c c }
\toprule
\textbf{Experiment} & \textbf{Vessel angle} & \textbf{Shunt outer diameter} & \textbf{Successes} & \textbf{Attempts} & \textbf{Success Rate} & \textbf{Avg trial time} & \multicolumn{2}{c}{\textbf{Failure Modes}} \\ 
\cmidrule{8-9}
& \textit{(deg)} & \textit{(mm)} & & & \textit{(\%)} & \textit{(s)} & (D) & (S) \\
\midrule
\multirow{2}*{No Dilation} & \multirow{2}*{0} & 8 & 20 & 20 & \textbf{100} & 9.0 & 0 & 0 \\
& & 14 & 0 & 20 & 0 & 10.1 & 0 & 20 \\
\midrule
Dilation Only & \multirow{2}*{0} & 8 & 19 & 20 & \textbf{95} & 13.7 & 1 & 0 \\
(No Screw Motion) & & 14 & 1 & 20 & 5 & 13.9 & 0 & 19 \\
\midrule
& \multirow{2}*{0} & 8 & 19 & 20 & \textbf{95} & 14.5 & 0 & 1 \\
& & 14 & 16 & 20 & \textbf{80} & 14.4 & 0 & 4 \\
\cmidrule{2-9}
Dilation + & \multirow{2}*{15} & 8 & 18 & 20 & \textbf{90} & 15.0 & 1 & 1 \\
Screw Motion & & 14 & 18 & 20 & \textbf{90} & 14.7 & 0 & 2 \\
\cmidrule{2-9}
& \multirow{2}*{30} & 8 & 17 & 20 & \textbf{85} & 13.8 & 1 & 2 \\
& & 14 & 10 & 20 & 50 & 14.8 & 0 & 10 \\
\bottomrule
\end{tabular}
}
\caption{\textbf{Experiment Results:} Success rate and mean trial time for shunt insertion with varying shunt with and without using dilation and screw motion, and including a vessel angle ablation. We track two failure modes: (D) dilation failure and (S) shunt insertion failure. 
}
\vspace{10pt}
\label{table:experiments}
\end{table*}

\subsection{Experimental Setup}
We perform experiments using the da Vinci Research Kit (dVRK) surgical robot with two cable-driven patient-side manipulator (PSM) arms \cite{jhudvrk2016}. The perception setup uses a Zivid OnePlus S camera mounted $0.5$\,m above the workspace with roughly a $50^{\circ}$ vertical incline, which captures RGBD images at 1920x1200 resolution. At the beginning of each trial, two mounted arms are used to fix two points $120^{\circ}$ apart on the rim of the vessel.

Our vessel phantom is constructed using the tube-top of a latex balloon with an inner diameter of 15mm (Fig. \ref{fig:trainingdata}) and a rim thickness of 1.5mm. The latex material is flexible, allowing the dVRK to stretch the rim for shunt insertion. The two shunt analogs are clear PVC vinyl tubes with 8mm and 14mm outer diameters, respectively. The shunts are flexible enough such that the dVRK can grasp them, but rigid enough to minimize deformation when they come in contact with the phantom. The plastic shunts we used are similar in stretchability and flexibility to—but roughly 1.5–2 times larger in diameter than—the MVP™ microvascular shunt (Covidien) (5.3mm-6.5mm) ~\citep{boudjemline2017covidien} used in clinical vascular shunt surgery. The recommended vessel sizes for the MVP™ microvascular shunt~\citep{boudjemline2017covidien} are between 3.0mm-5.0mm, resulting in the outer diameter of the shunt being larger than the diameter of the recommended vessel by 1.5mm-2.3mm. The smallest difference between the diameters of our shunts and vessel phantom is 1mm.  
\subsection{Metrics and Failure Modes}
Recall that we consider a trial a success if the rim of the shunt is fully enclosed by the vessel after both PSMs release their grips and retract back to their original position. For each trial, we record the success or failure of the attempt, along with the elapsed time. We classify each failure as one of the following failure modes:

\subsubsection{Dilation Failure (D)}
When attempting to dilate the vessel, the robot fails to grasp the vessel phantom rim or fails to pull outward. This happens when the circle fitting experiences noisy depth data, the robot's calibration is incorrect, or the rim is irregularly shaped, causing the planned grasp point to be located off of the actual rim.

\subsubsection{Shunt Insertion Failure (S)}
After executing the shunt insertion procedure, the complete rim of the shunt is not enclosed by the vessel phantom. This occurs when the shunt completely misses the vessel phantom or when any part of the shunt's rim is sticking out above the top of the vessel phantom.

\subsection{Full Pipeline}
We perform 20 trials with the 8mm shunt using the full pipeline, as described in Section \ref{sec:method}, and obtain a success rate of 95\% with an average trial time of 13.7s. We encountered 1 dilation failure (D), and  0 shunt insertion failures (S). We performed 20 trials with the 14mm shunt using the full pipeline and obtained a success rate of 80\% with an average trial time of 14.4s. We encountered 0 dilation failures (D), and 4 shunt insertion failures (S). We report full results in Table~\ref{table:experiments}.

\subsection{Baselines}
We compare our method against two baselines to test the impact of different parts of our pipeline. We report full results in Table~\ref{table:experiments}.

\subsubsection{No Dilation}
To evaluate the effect of the dilation step on the shunt insertion, we execute the pipeline with all steps except for the vessel phantom grasping and dilation.

We perform 20 trials without dilation with both the 8mm and 14mm shunts, and report the success rates as 100\% and 0\% with average trial times of 9.0s and 10.1s respectively. 

In the smaller 8mm outer diameter shunt case, the shunt was smaller compared to the vessel that—in general—the shunt did not contact the sides of the vessel phantom near the rim as it moved downward, resulting in a success rate that was not negatively impacted. Conversely, there were no successful no-dilation insertions of the larger 14mm shunt, as when attempting to insert it, the sections of the vessel phantom's rim between the grasping points buckled downward, causing the end of the shunt to rest partially outside of the vessel phantom.

\begin{figure}
\includegraphics[width=1.0\linewidth]{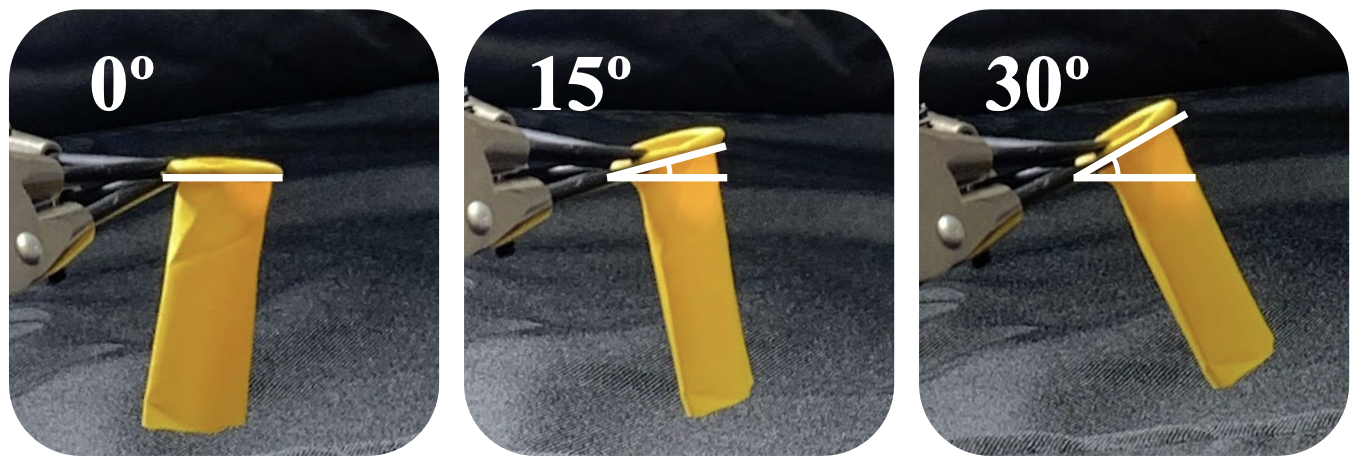}
\centering
\caption{\textbf{Angles of ablation.} We evaluate our pipeline using phantoms grasped at increasingly adversarial angles, performing trials at 0º, 15º, and 30º from the horizontal.}
\label{fig:ablation_angles}
    \vspace{5pt}
\end{figure}

\subsubsection{No Screw Motion} 
To evaluate the effect of the screw motion at the end of the shunt insertion step, we perform trials where no screw motion is used. Instead, after performing the chamfer tilt insertion and then straightening out the shunt, the robot proceeds to release the grippers.

We perform 20 trials of this baseline with both the 8mm and 14mm shunts, with average trial times of 13.7s and 13.9s, and success rates of 95\% and 5\%, respectively. We report full results in Table~\ref{table:experiments}. The lower recorded trial durations result from the omission of the final screw motion; however, this faster execution comes at the expense of the success rate for the large shunt, where the low tolerance between the shunt and vessel causes the edge of the shunt to get stuck on the outside the rim when the screw motion is omitted, demonstrated by the increase in shunt insertion failures. When the inner diameter of the rim of the vessel phantom is too close to the shunt's outer diameter, a portion of the circumference of the shunt's opening may be placed outside of the vessel phantom, a situation that otherwise would be corrected by the screw motion's rotation.

\subsubsection{Vessel Phantom Angle Ablation}
We also perform an ablation study of the method by varying the angle between the rim of the vessel phantom and the ground. We evaluate the full pipeline by rotating the vessel phantom at 2 different angles: 15$^{\circ}$ and 30$^{\circ}$ with respect to the workspace frame (Fig.~\ref{fig:ablation_angles}).

We perform 20 trials per ablation stage, and report results in Table~\ref{table:experiments}. For the 8mm shunt, the success rates for the 15$^{\circ}$ and 30$^{\circ}$ vessel phantom rim angles are 90\% and 85\% with average trial times of 15.0s and 13.8s respectively.

For the 14mm outer diameter shunt, the success rates for the 15$^{\circ}$ and 30$^{\circ}$ vessel phantom rim angles are 90\% and 50\% respectively. The negative impact on the success rate between the low-angle (0$^{\circ}$, 15$^{\circ}$) and high-angle cases (30$^{\circ}$), results from a decrease in the effectiveness of the initial chamfer tilt insertion; at high angles, the initial chamfer tilt insertion results in an orientation that is not fully seated within the vessel rim, and even with the screw motion, one part of the shunt ends up outside of the vessel phantom. On the other hand, in both the 0$^{\circ}$ and 15$^{\circ}$ cases, the larger shunt was initially inserted at a well-grounded position for the following screw motion.

\subsubsection{Segmentation Model Color Sensitivity}
To evaluate the independence of our detection model from our specific colored vessel phantoms, we train a version of our segmentation U-Net on grayscale versions of our input images. We use a separate set of $3200$ images for training the grayscale model, and grayscale versions of the same $805$ validation images for evaluating the models. We compare the outputs of the RGB segmentation model and the grayscale segmentation model to the ground truth labels using their Intersection over Union (IoU). The RGB input segmentation model had an average IoU of $0.62$ on the validation set, and the grayscale input segmentation model produced an average IoU of $0.58$. As the measured performance of the two models on our validation set was nearly equivalent, we found no reason to suspect that our detection system would fail to converge under more difficult color conditions, such as those faced during an in-vivo operation. However, we hope that future work will be able to put this assumption to the test physically.
\section{Limitations}
The approach and physical environment we consider have notable limitations: The vessel phantom and shunts used are scaled up by $50$—$100\%$ from their clinical use cases, and their appearance may not align with that of tissue during a real operation. The vessel phantom is held by two stationary grippers, but these two points may not remain stationary in a real clinical setting, since a human holds them. To account for this, future work could include visual servoing algorithms to iteratively move the gripper to a correct grasp position. Furthermore, the shunt placement in one of the grippers may not be known ahead of time in a real scenario. The stress applied to the vessel when inserting the shunt is not explicitly considered and accounted for by the algorithm. The initial shunt insertion point is manually tuned relative to the vessel phantom's center for the given environment.
\section{Discussion}

In this paper, we formulate the Automated Vascular Shunt Insertion problem and present an algorithmic pipeline designed to perform it on a da Vinci Research Kit. Our results show that the method proposed in this paper achieves a high success rate even with tight tolerances and varying vessel orientations. To the best of our knowledge, this work is the first to study and implement a method for automating vascular shunting surgery. 

In future work, we will consider using more clinically realistic materials and sizes for the vessel and the shunt. Furthermore, implementing and evaluating automated vascular shunt insertion pipelines on robots such as the SRI Taurus~\cite{taurus} can more closely resemble hardware deployed in battlefield scenarios. To accommodate for smaller shunt and vessel phantom cases and improving the success rate of the current pipeline, future work may explore applying irregularly-shaped rim fitting to loosen the assumption of a circular rim and utilize visual servoing for both grasping the vessel phantom and inserting the shunt.

Future work may also consist of automating other subtasks of vascular shunt insertion, such as adding a ligature knot to stabilize the shunt position, and using two other robotic arms to perform the initial grasping on the vessel. Parts of the automated vascular shunt insertion algorithm can be extended and applied to similar procedures, such as carotid endarterectomy~\cite{uno2020carotid}.
\section*{Acknowledgments}
\textcolor{black}{\small{This research was performed at the AUTOLAB at UC Berkeley in affiliation with the Berkeley AI Research (BAIR) Lab and the CITRIS "People and Robots" (CPAR) Initiative. This work is supported in part by the Technology \& Advanced Telemedicine Research Center (TATRC) project W81XWH-19-C-0096 under a medical Telerobotic Operative Network (TRON) project led by SRI International and donations from Intuitive Surgical. The da Vinci Research Kit is supported by the National Science Foundation, via the National Robotics Initiative (NRI) \cite{jhudvrk2016}. We thank Jules Dedieu and Jonathan Pei for providing helpful information and Roy Lin, Zach Tam, and Varun Kamat for their useful feedback.}}

\renewcommand*{\bibfont}{\footnotesize}
\printbibliography

@String { icra    = {IEEE International Conference on Robotics and Automation (ICRA)} }

@String { ieeera  = {IEEE Robotics and Automation Letters (RA-L)} }

@String { ismr    = {International Symposium on Medical Robotics (ISMR)} }

@article{subramanian2008decade,
  title={A decade's experience with temporary intravascular shunts at a civilian level I trauma center},
  author={Subramanian, Anuradha and Vercruysse, Gary and Dente, Christopher and Wyrzykowski, Amy and King, Erin and Feliciano, David V},
  journal={Journal of Trauma and Acute Care Surgery},
  volume={65},
  number={2},
  pages={316--326},
  year={2008},
  publisher={LWW}
}

@article{liu2021clinical,
  title={The clinical application of robot-assisted ventriculoperitoneal shunting in the treatment of hydrocephalus},
  author={Liu, De-feng and Liu, Huan-guang and Zhang, Kai and Meng, Fan-gang and Yang, An-chao and Zhang, Jian-guo},
  journal={Frontiers in Neuroscience},
  pages={971},
  year={2021},
  publisher={Frontiers}
}

@article{doddamani2021robot,
  title={Robot-guided ventriculoperitoneal shunt in slit-like ventricles},
  author={Doddamani, Ramesh S and Meena, Rajesh and Sawarkar, Dattaraj and Singh, Pankaj and Agrawal, Deepak and Singh, Manmohan and Chandra, Poodipedi S and others},
  journal={Neurology India},
  volume={69},
  number={2},
  pages={446},
  year={2021},
  publisher={Medknow Publications}
}

@article{thananjeyan2022all,
  title={All You Need is LUV: Unsupervised Collection of Labeled Images using Invisible UV Fluorescent Indicators},
  author={Thananjeyan, Brijen and Kerr, Justin and Huang, Huang and Gonzalez, Joseph E and Goldberg, Ken},
  journal={arXiv preprint arXiv:2203.04566},
  year={2022}
}

@misc{cannon_2018, title={Shunt procedure}, url={https://www.hopkinsmedicine.org/neurology_neurosurgery/centers_clinics/cerebral-fluid/procedures/shunts.html}, journal={Johns Hopkins Medicine in Baltimore, MD}, author={Cannon, Julia}, year={2018}, month={Feb}}

@article{Voiglio2016AbbreviatedLO,
  title={Abbreviated laparotomy or damage control laparotomy: Why, when and how to do it?},
  author={Eric J. Voiglio and Vincent Dubuisson and Damien Massalou and Yoann Baudoin and Jeanne Caillot and Christian L{\'e}toublon and Catherine Arvieux},
  journal={Journal of visceral surgery},
  year={2016},
  volume={153 4 Suppl},
  pages={13-24}
}

@inproceedings{paradis2020intermittent,
  booktitle={{Intermittent Visual Servoing: Efficiently Learning Policies Robust to Instrument Changes for High-precision Surgical Manipulation}},
  author={Paradis, Samuel and Hwang, Minho and Thananjeyan, Brijen and Ichnowski, Jeffrey and Seita, Daniel and Fer, Danyal and Low, Thomas and Gonzalez, Joseph E and Goldberg, Ken},
  publisher=icra,
  year={2021}
}

@inproceedings{saeidi_suturing_icra_2019,
  author = {Saeidi, H and Le, H N D and Opfermann, J D and Leonard, S and Kim, A and Hsieh, M H and Kang, J U and Krieger, A},
  booktitle = {{Autonomous Laparoscopic Robotic Suturing with a Novel Actuated Suturing Tool and 3D Endoscope}},
  publisher = icra,
  Year = {2019}
}

@inproceedings{rosen_icra_tissues_2019,
  author = {Changyeob Shin and Peter Walker Ferguson and Sahba Aghajani Pedram and Ji Ma and Erik P. Dutson and Jacob Rosen},
  booktitle = {{Autonomous Tissue Manipulation via Surgical Robot Using Learning Based Model Predictive Control}},
  publisher = icra,
  Year = {2019}
}

@inproceedings{seita_icra_2018,
  author = {Daniel Seita and Sanjay Krishnan and Roy Fox and Stephen McKinley and John Canny and Kenneth Goldberg},
  booktitle = {{Fast and Reliable Autonomous Surgical Debridement with Cable-Driven Robots Using a Two-Phase Calibration Procedure}},
  publisher = icra,
  Year = {2018}
}

@inproceedings{thananjeyan2017multilateral,
  booktitle={{Multilateral Surgical Pattern Cutting in 2D Orthotropic Gauze with Deep Reinforcement Learning Policies for Tensioning}},
  author={Thananjeyan, Brijen and Garg, Animesh and Krishnan, Sanjay and Chen, Carolyn and Miller, Lauren and Goldberg, Ken},
  publisher=icra,
  year={2017},
}

@inproceedings{sen2016automating,
   Author = {Sen, S. and Garg, A. and Gealy, D. V. and McKinley, S. and Jen, Y. and Goldberg, K.},
   booktitle = {{Automating Multiple-Throw Multilateral Surgical Suturing with a Mechanical Needle Guide and Sequential Convex Optimization}},
   publisher = icra,
   Year = {2016}
 }

@inproceedings{murali2015learning,
  booktitle={{Learning by Observation for Surgical Subtasks: Multilateral Cutting of 3D Viscoelastic and 2D Orthotropic Tissue Phantoms}},
  author={Murali, Adithyavairavan and Sen, Siddarth and Kehoe, Ben and Garg, Animesh and McFarland, Seth and Patil, Sachin and Boyd, W Douglas and Lim, Susan and Abbeel, Pieter and Goldberg, Ken},
  publisher = icra,
  year={2015},
}

@article{asymmetrictumor,
author = {Rosas Gonzalez, Sarahí and Birgui-Sekou, Taibou and Hidane, Moncef and Zemmoura, Ilyess},
year = {2021},
month = {09},
pages = {609646},
title = {Asymmetric Ensemble of Asymmetric U-Net Models for Brain Tumor Segmentation With Uncertainty Estimation},
volume = {12},
journal = {Frontiers in Neurology},
doi = {10.3389/fneur.2021.609646}
}

@InProceedings{Kehoe2014,
   booktitle = {{Autonomous Multilateral Debridement with the Raven Surgical Robot}},
   Author= {Kehoe, B. and Kahn, G. and Mahler, J. and Kim, J. and Lee, A. and Lee, A. and Nakagawa, K. and Patil, S. and Boyd, W.D. and Abbeel, P. and Goldberg, K.},
   publisher = icra,
   Year= {2014}
 }

@inproceedings{automated_needle_pickup_2018,
  booktitle = {{Automated Pick-up of Suturing Needles for Robotic Surgical Assistance}},
  Author = {D’Ettorre, C and Dwyer, G and Du, X and Chadebecq, F and  Vasconcelos, F and De Momi, E and Stoyanov, D},
  publisher = icra,
  Year = {2018}
}

@inproceedings{hwang2020efficiently,
  booktitle={{Efficiently Calibrating Cable-Driven Surgical Robots With RGBD Fiducial Sensing and Recurrent Neural Networks}},
  author={Hwang, Minho and Thananjeyan, Brijen and Paradis, Samuel and Seita, Daniel and Ichnowski, Jeffrey and Fer, Danyal and Low, Thomas and Goldberg, Ken},
  publisher=ieeera,
  year={2020}
}

@inproceedings{hwang2020applying,
  author = {Minho Hwang and Daniel Seita and Brijen Thananjeyan and Jeffrey Ichnowski and Samuel Paradis and Danyal Fer and Thomas Low and Ken Goldberg},
  booktitle = {{Applying Depth-Sensing to Automated Surgical Manipulation with a da Vinci Robot}},
  publisher = ismr,
  Year = {2020}
}

@inproceedings{auto_peg_transfer_2015,
  booktitle = {{Autonomous Operation in Surgical Robotics}},
  author = {Jacob Rosen and Ji Ma},
  publisher = {Mechanical Engineering},
  volume={137},
  number={9},
  year = 2015,
}

@article{bamba2021object,
  title={Object and anatomical feature recognition in surgical video images based on a convolutional neural network},
  author={Bamba, Yoshiko and Ogawa, Shimpei and Itabashi, Michio and Shindo, Hironari and Kameoka, Shingo and Okamoto, Takahiro and Yamamoto, Masakazu},
  journal={International Journal of Computer Assisted Radiology and Surgery},
  volume={16},
  number={11},
  pages={2045--2054},
  year={2021},
  publisher={Springer}
}

@article{sanchez2022gauze,
  title={Gauze Detection and Segmentation in Minimally Invasive Surgery Video Using Convolutional Neural Networks},
  author={S{\'a}nchez-Brizuela, Guillermo and Santos-Criado, Francisco-Javier and Sanz-Gobernado, Daniel and de la Fuente-L{\'o}pez, Eusebio and Fraile, Juan-Carlos and P{\'e}rez-Turiel, Javier and Cisnal, Ana},
  journal={Sensors},
  volume={22},
  number={14},
  pages={5180},
  year={2022},
  publisher={Multidisciplinary Digital Publishing Institute}
}

@article{yang2020image,
  title={Image-based laparoscopic tool detection and tracking using convolutional neural networks: a review of the literature},
  author={Yang, Congmin and Zhao, Zijian and Hu, Sanyuan},
  journal={Computer Assisted Surgery},
  volume={25},
  number={1},
  pages={15--28},
  year={2020},
  publisher={Taylor \& Francis}
}

@article{bamba2021automated,
  title={Automated recognition of objects and types of forceps in surgical images using deep learning},
  author={Bamba, Yoshiko and Ogawa, Shimpei and Itabashi, Michio and Kameoka, Shingo and Okamoto, Takahiro and Yamamoto, Masakazu},
  journal={Scientific Reports},
  volume={11},
  number={1},
  pages={1--8},
  year={2021},
  publisher={Nature Publishing Group}
}

@article{liu2022real,
  title={Real-Time Surgical Tool Detection in Computer-aided Surgery Based on Enhanced Feature Fusion Convolutional Neural Network},
  author={Liu, Kaidi and Zhao, Zijian and Shi, Pan and Li, Feng and Song, He},
  journal={Journal of Computational Design and Engineering},
  year={2022}
}

@inproceedings{boonkong2022surgical,
  title={Surgical Instrument Detection for Laparoscopic Surgery using Deep Learning},
  author={Boonkong, Apiwat and Hormdee, Daranee and Sonsilphong, Suphachoke and Khampitak, Kovit},
  booktitle={2022 19th International Conference on Electrical Engineering/Electronics, Computer, Telecommunications and Information Technology (ECTI-CON)},
  pages={1--4},
  year={2022},
  organization={IEEE}
}

@article{gong2021using,
  title={Using deep learning to identify the recurrent laryngeal nerve during thyroidectomy},
  author={Gong, Julia and Holsinger, F Christopher and Noel, Julia E and Mitani, Sohei and Jopling, Jeff and Bedi, Nikita and Koh, Yoon Woo and Orloff, Lisa A and Cernea, Claudio R and Yeung, Serena},
  journal={Scientific Reports},
  volume={11},
  number={1},
  pages={1--11},
  year={2021},
  publisher={Nature Publishing Group}
}

@article{goto2022image,
  title={Image-guided surgery with a new tumour-targeting probe improves the identification of positive margins},
  author={Goto, Masahide and Ryoo, Ingeun and Naffouje, Samer and Mander, Sunam and Christov, Konstantin and Wang, Jing and Green, Albert and Shilkaitis, Anne and Gupta, Tapas K Das and Yamada, Tohru},
  journal={EBioMedicine},
  volume={76},
  pages={103850},
  year={2022},
  publisher={Elsevier}
}

@inproceedings{chheda2020gastrointestinal,
  title={Gastrointestinal tract anomaly detection from endoscopic videos using object detection approach},
  author={Chheda, Tejas and Iyer, Rithvika and Koppaka, Soumya and Kalbande, Dhananjay},
  booktitle={International Symposium on Visual Computing},
  pages={494--505},
  year={2020},
  organization={Springer}
}

@article{li2021semi,
  title={A semi-supervised deep learning approach for circular hole detection on composite parts},
  author={Li, Guanhua and Yang, Shuang and Cao, Siming and Zhu, Weidong and Ke, Yinglin},
  journal={The Visual Computer},
  volume={37},
  number={3},
  pages={433--445},
  year={2021},
  publisher={Springer}
}

@article{ling4075794deep,
  title={A Deep Learning Approach for Plate Circular Hole Detection on Composite Parts},
  author={Ling, Zihao and Gan, Zhong and Shi, Wangxing and Yang, Le and Ma, Boyu and Xue, Chao},
  journal={Available at SSRN 4075794}
}

@article{prabuwono2019automated,
  title={Automated visual inspection for bottle caps using fuzzy logic},
  author={Prabuwono, Anton Satria and Usino, Wendi and Yazdi, Leila and Basori, Ahmad Hoirul and Bramantoro, Arif and Syamsuddin, Irfan and Yunianta, Arda and Allehaibi, Khalid Hamed S},
  journal={TEM Journal},
  volume={8},
  number={1},
  pages={107},
  year={2019},
  publisher={UIKTEN-Association for Information Communication Technology Education and~…}
}

@inproceedings{ektesabi2011exact,
  title={Exact pupil and iris boundary detection},
  author={Ektesabi, A and Kapoor, A},
  booktitle={The 2nd International Conference on Control, Instrumentation and Automation},
  pages={1217--1221},
  year={2011},
  organization={IEEE}
}

@article{krishnan2019swirl,
  title={SWIRL: A sequential windowed inverse reinforcement learning algorithm for robot tasks with delayed rewards},
  author={Krishnan, Sanjay and Garg, Animesh and Liaw, Richard and Thananjeyan, Brijen and Miller, Lauren and Pokorny, Florian T and Goldberg, Ken},
  journal={The International Journal of Robotics Research},
  volume={38},
  number={2-3},
  pages={126--145},
  year={2019},
  publisher={SAGE Publications Sage UK: London, England}
}

@article{wilcox2021learning,
  title={Learning to Localize, Grasp, and Hand Over Unmodified Surgical Needles},
  author={Wilcox, Albert and Kerr, Justin and Thananjeyan, Brijen and Ichnowski, Jeffrey and Hwang, Minho and Paradis, Samuel and Fer, Danyal and Goldberg, Ken},
  journal={arXiv preprint arXiv:2112.04071},
  year={2021}
}

@misc{taurus,
  title = {Taurus: This small robot is reaching new heights and solving once-thought impossible challenges},
  howpublished = {\url{https://medium.com/dish/taurus-this-small-robot-is-reaching-new-heights-and-solving-once-thought-impossible-challenges-e858fdbbb4ab}},
%   note = {Accessed: 2022-02-18}
}

@inproceedings{ronneberger2015u,
  title={U-net: Convolutional networks for biomedical image segmentation},
  author={Ronneberger, Olaf and Fischer, Philipp and Brox, Thomas},
  booktitle={International Conference on Medical image computing and computer-assisted intervention},
  pages={234--241},
  year={2015},
  organization={Springer}
}

@article{saeidi2022autonomous,
  title={Autonomous robotic laparoscopic surgery for intestinal anastomosis},
  author={Saeidi, Hamed and Opfermann, Justin D and Kam, Michael and Wei, Shuwen and L{\'e}onard, Simon and Hsieh, Michael H and Kang, Jin U and Krieger, Axel},
  journal={Science Robotics},
  volume={7},
  number={62},
  pages={eabj2908},
  year={2022},
  publisher={American Association for the Advancement of Science}
}

@article{rasmussen2006use,
  title={The use of temporary vascular shunts as a damage control adjunct in the management of wartime vascular injury},
  author={Rasmussen, Todd E and Clouse, W Darrin and Jenkins, Donald H and Peck, Michael A and Eliason, Jonathan L and Smith, David L},
  journal={Journal of Trauma and Acute Care Surgery},
  volume={61},
  number={1},
  pages={8--15},
  year={2006},
  publisher={LWW}
}

@article{garcia2009trauma,
  title={Trauma Pod: a semi-automated telerobotic surgical system},
  author={Garcia, Pablo and Rosen, Jacob and Kapoor, Chetan and Noakes, Mark and Elbert, Greg and Treat, Michael and Ganous, Tim and Hanson, Matt and Manak, Joe and Hasser, Chris and others},
  journal={The International Journal of Medical Robotics and Computer Assisted Surgery},
  volume={5},
  number={2},
  pages={136--146},
  year={2009},
  publisher={Wiley Online Library}
}

@article{eger1971use,
  title={The use of a temporary shunt in the management of arterial vascular injuries},
  author={Eger, M and Golcman, L and Goldstein, A and Hirsch, M},
  journal={Surgery, gynecology \& obstetrics},
  volume={132},
  number={1},
  pages={67--70},
  year={1971}
}

@article{malan1963physio,
  title={Physio-and anatomo-pathology of acute ischemia of the extremities},
  author={Malan, E and Tattoni, G},
  journal={The Journal of cardiovascular surgery},
  volume={4},
  pages={212--225},
  year={1963}
}

@article{burkhardt2010large,
  title={A large animal survival model (Sus scrofa) of extremity ischemia/reperfusion and neuromuscular outcomes assessment: a pilot study},
  author={Burkhardt, Gabriel E and Spencer, Jerry R and Gifford, Shaun M and Propper, Brandon and Jones, Lyell and Sumner, Nathan and Cowart, Jerry and Rasmussen, Todd E},
  journal={Journal of Trauma and Acute Care Surgery},
  volume={69},
  number={1},
  pages={S146--S153},
  year={2010},
  publisher={LWW}
}

@article{rasmussen2006echelons,
  title={Echelons of care and the management of wartime vascular injury: a report from the 332nd EMDG/Air Force Theater Hospital, Balad Air Base, Iraq},
  author={Rasmussen, Todd E and Clouse, W Darrin and Jenkins, Donald H and Peck, Michael A and Eliason, Jonathan L and Smith, David L},
  journal={Perspectives in vascular surgery and endovascular therapy},
  volume={18},
  number={2},
  pages={91--99},
  year={2006},
  publisher={Sage Publications Sage CA: Thousand Oaks, CA}
}

@article{jhudvrk2016,
title={Software Framework for Research in Semi-Autonomous Teleoperation},
author={Taylor, Russell H. and Kazanzides, Peter},
year={2016}}

@article{fischler1981random,
  title={Random sample consensus: a paradigm for model fitting with applications to image analysis and automated cartography},
  author={Fischler, Martin A and Bolles, Robert C},
  journal={Communications of the ACM},
  volume={24},
  number={6},
  pages={381--395},
  year={1981},
  publisher={ACM New York, NY, USA}
}

@article{boudjemline2017covidien,
  title={Covidien micro vascular plug in congenital heart diseases and vascular anomalies: a new kid on the block for premature babies and older patients},
  author={Boudjemline, Younes},
  journal={Catheterization and Cardiovascular Interventions},
  volume={89},
  number={1},
  pages={114--119},
  year={2017},
  publisher={Wiley Online Library}
}

@article{uno2020carotid, 
  title={Surgical Technique for Carotid Endarterectomy: Current Methods and Problems},
  author={Uno, Masaaki and Takai, Hiroki and Yagi, Kenji and Matsubara Shunji},
  journal={Neurol Med Chir (Tokyo)},
  year={2020},
  volume={60}
}
\clearpage

\end{document}